\documentclass{llncs}
\usepackage{makeidx}  % allows for indexgeneration

\usepackage{epsfig}
\usepackage{subfigure}
\usepackage{graphicx}
\usepackage{amsmath}
\usepackage{amssymb}
\usepackage{algpseudocode}
\usepackage{algorithm}
\usepackage{color}

\begin{document}

%\frontmatter          % for the preliminaries
%
\pagestyle{headings}  % switches on printing of running heads
%\addtocmark{Hamiltonian Mechanics} % additional mark in the TOC
%
%
\mainmatter              % start of the contributions
\title{Recurrent Fully Convolutional Neural Networks for Multi-slice MRI Cardiac Segmentation}

%\author{Anonymous Authors}
\author{Rudra P K Poudel, Pablo Lamata \and Giovanni Montana}

\institute{Department of Biomedical Engineering, King's College London, SE1 7EH, UK} 
%\institute{*} 

\maketitle

\pagenumbering{gobble}

%
% Modify the bibliography environment to call for the author-year
% system. This is done normally with the citeauthoryear option
% for a particular contribution.
%\makeatletter
%\renewenvironment{thebibliography}[1]
%     {\section*{\refname}
%      \small
%      \list{}%
%           {\settowidth\labelwidth{}%
%            \leftmargin\parindent
%            \itemindent=-\parindent
%            \labelsep=\z@
%            \if@openbib
%              \advance\leftmargin\bibindent
%              \itemindent -\bibindent
%              \listparindent \itemindent
%              \parsep \z@
%            \fi
%            \usecounter{enumiv}%
%            \let\p@enumiv\@empty
%            \renewcommand\theenumiv{}}%
%      \if@openbib
%        \renewcommand\newblock{\par}%
%      \else
%        \renewcommand\newblock{\hskip .11em \@plus.33em \@minus.07em}%
%      \fi
%      \sloppy\clubpenalty4000\widowpenalty4000%
%      \sfcode`\.=\@m}
%     {\def\@noitemerr
%       {\@latex@warning{Empty `thebibliography' environment}}%
%      \endlist}
%      \def\@cite#1{#1}%
%      \def\@lbibitem[#1]#2{\item[]\if@filesw
%        {\def\protect##1{\string ##1\space}\immediate
%      \write\@auxout{\string\bibcite{#2}{#1}}}\fi\ignorespaces}
%\makeatother
%
%------------------------------------------------------------------------
\begin{abstract}
In cardiac magnetic resonance imaging, fully-automatic segmentation of the heart enables precise structural and functional measurements to be taken, e.g. from short-axis MR images of the left-ventricle. In this work we propose a recurrent fully-convolutional network (RFCN) that learns image representations from the full stack of 2D slices and has the ability to leverage inter-slice spatial dependences through internal memory units. RFCN combines anatomical detection and segmentation into a single architecture that is trained end-to-end thus significantly reducing computational time, simplifying the segmentation pipeline, and potentially enabling real-time applications. We report on an investigation of RFCN using two datasets, including the publicly available MICCAI 2009 Challenge dataset. Comparisons have been carried out between fully convolutional networks and deep restricted Boltzmann machines, including a recurrent version that leverages inter-slice spatial correlation. Our studies suggest that RFCN produces state-of-the-art results and can substantially improve the delineation of contours near the apex of the heart. 

\end{abstract}

%
%------------------------------------------------------------------------
%------------------------------------------------------------------------
\section{Introduction}
% no \IEEEPARstart
Cardiovascular disease is one of the major causes of death in the world. Physicians use imaging technologies such as magnetic resonance imaging (MRI) to estimate structural (e.g. volume) and functional (e.g. ejection fraction) cardiac parameters for both diagnosis and disease management. Fully-automated estimation of such parameters can facilitate early diagnosis of the disease and has the potential to remove the more mechanistic aspects of a radiologist's assessment. As such, lately there has been increasing interest in machine learning algorithms for fully automatic left-ventricle (LV) segmentation \cite{avendi2016,ngo2014,hu2013,huang2011,jolly2009}. This is a challenging task due to the variability of LV shape across slices, cardiac phases, patients and scanning machines as well as weak boundaries of LV due to the presence of blood flow, papillary muscles and trabeculations. A review of LV segmentation methods in short-axis cardiac MR images can be found in \cite{petitjean2011}. 

The main image analysis approaches to LV segmentation can be grouped into three broad categories: active contour models, machine learning models, and hybrid versions that combine elements of the two approaches. Active contour models with either explicit \cite{kass1988} or implicit \cite{li2010} contour representations minimize an energy function composed of internal and external constraints. The internal constraints represent continuity and smoothness of the contour and external constraints represent appearance and shape of the target object. However, designing appropriate energy functions that can handle all sources of variability is challenging. Also, the quality of the segmentations produced by these methods typically depends on the region-of-interest (ROI) used to initialise the algorithms. Machine learning approaches have been proposed to circumvent some of these issues  \cite{huang2004,ngo2013,ngo2014,avendi2016} at the expense of collecting large training datasets with a sufficient number of examples. Investigating hybrid methods that combine some elements of both approaches is an active research area \cite{georgescu2005}. %Nevertheless, Markovian models are complex in general and require hand tuning of lots of parameters. {\bf MARKOV MODELS ARE NOT 'MACHINE LEARNING'}

Current state-of-the-art LV segmentation approaches rely on deep artificial neural networks \cite{ngo2013,ngo2014,avendi2016}. Typically, these solutions consists of three distinct stages carried out sequentially. Initially, the LV is localised within each two-dimensional slice; then the LV is segmented, and finally the segmentation is further refined to improve its quality. For instance, a pipeline consisting of Deep Belief Networks (DBNs) for both localisation and segmentation, followed by a level-set methodology, has shown to generate high-quality segmentations \cite{ngo2014}. In more recent work, a different pipeline has been proposed that consists of convolutional neural networks for initial LV detection, followed by a segmentation step deploying stacked autoencoders, and a fine-tuning strategy also based on level-sets methodology \cite{avendi2016}. The latter approach has been proved to produce state-of-the-art results on the MICCAI 2009 LV segmentation challenge \cite{radau2009}. Both approaches share a number of common features. First, the segmentation is carried out using two-dimensional patches that are independently extracted from each MRI slice. Second, they use a separate architecture for the two tasks, localization and segmentation. Third, different neural network architectures are trained for cardiac MR slices containing the base and apex of the heart, due to the observed heterogeneity in local shape variability. 

In this work we investigate a neural network architecture, trained end-to-end, that learns to detect and segment the LV jointly from the the entire stack of short-axis images rather than operating on individual slices. Recently, fully convolutional networks (FCN) have been proposed for the segmentation of 2D images \cite{long2015}. They take arbitrarily sized input images, and use feature pooling coupled with an upsampling step to produce same size outputs delivering the segmentation. Compared to more traditional sliding-window approaches, FCNs are more efficient. They have received increasing interest lately as they unify object localization and segmentation in a single process by extracting both global and local context effectively \cite{long2015,ronneberger2015}. Applications of FCNs to medical imaging segmentation problems have also started to appear, for instance for the identification of neuronal structures in electron microscopic recordings \cite{ronneberger2015}. In independent work, Valipour et al. \cite{valipour2016} have recently adapted recurrent fully convolutional networks for video segmentation.

Here we propose an extension of FCNs, called Recurrent Fully-Convolutional Networks (RFCN), to directly address the segmentation problem in multi-slice MR images. We are motivated by the desire to exploit the spatial dependences that are observed across adjacent slices and learn image features that capture the global anatomical structure of the heart from the full image stack. We investigate whether exploiting this information is beneficial for accurate anatomical segmentation, especially for cardiac regions with weak boundaries, e.g. poor structural contrast due to the presence of blood flow, papillary muscles and trabeculations.

\section{Datasets} \label{datasets}

Our experiments are based on two independent datasets consisting of short-axis cardiac MR images for which the endocardium has been manually segmented by expert radiologists in each axial slice. Further details are provided below.

\subsection{MICCAI dataset}

The MICCAI 2009 LV Segmentation Challenge \cite{radau2009} dataset was made publicly available by the Sunnybrook Health Sciences Center (Canada) and has been extensively used to compare a number of LV segmentation algorithms \cite{jolly2009,huang2011,hu2013,ngo2013,ngo2014,avendi2016}. It consists of $45$ CINE MRI images from a number of different pathologies. The individual exams have been pre-grouped into training, validation and online testing subsets. Each subset contains $15$ cases of which $4$ heart failure with infarction (HF-I), $4$ heart failure without infarction (HF), $4$ LV hypertrophy (HYP) and $3$ healthy subjects. However, the clinical information has not been used by any of the algorithms discussed here and in our experiments. All the images were obtained during breath-hold sessions lasting $10-15$ seconds with a temporal resolution of $20$ cardiac phases over the heart cycle. A typical phase, end diastole (ED) or end systole (ES), contains $6-12$ short-axis slices obtained from the base to apex. In all the images, the slice thickness is $8$mm, the inter-slice gap is $8$mm, the field of view is $320$mm $\times$ $320$mm and the pixel size is $256$ $\times 256$. In all $45$ samples, LV endocardial contours were drawn by an experienced cardiologist by taking 2D slices at both the end-systolic and end-diastolic phases, and then independently confirmed by a second reader. The manual segmentations were used as ground truth for the evaluation of the proposed models. Each set consists of $30$ sequences ($15$ samples for each one of the two cardiac phases) with an average sequence length $8.94$ slices.

\begin{figure}[tb]
	\begin{center}
		\includegraphics[width=0.45\columnwidth, clip, trim= 0 8.95cm 16.5cm 0]{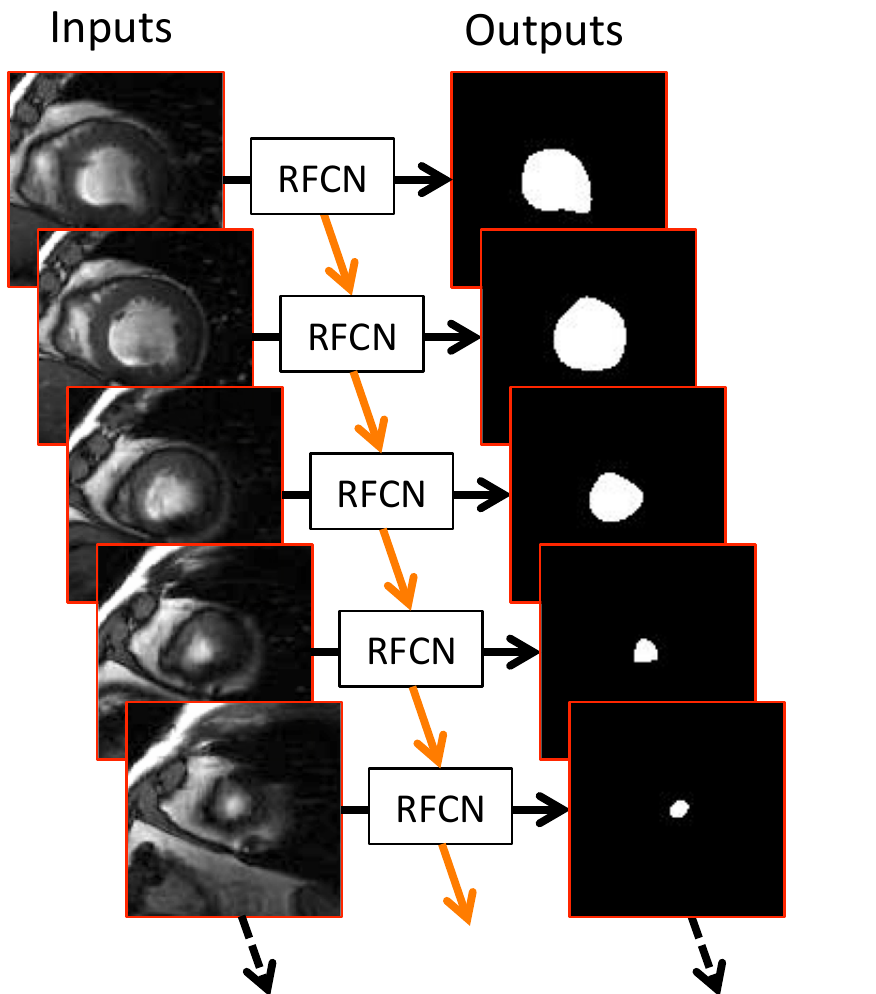}
	\end{center}
	\caption{A stack of short-axis cardiac MR slices (left) with corresponding left-ventricular binary masks (right) for a cardiac phase. The proposed RFCN leverages the spatial correlations that can be observed moving from the base of the heart to the apex.}
	\label{fig:rfcn_inputs}
\end{figure}

\subsection{PRETERM dataset}

A second and larger dataset was used for an independent evaluation of all the cardiac segmentation algorithms. The dataset consists of $234$ subjects used to study perinatal factors modifying the left ventricular parameter \cite{lewandowski2013}. All the individuals are between $20$ to $39$ years of age. Of these, $102$ were followed prospectively since preterm birth, and are characterised by an average gestational age of $30.3\pm 2.5$ weeks and a birth weight of $1.3 \pm 0.3$ kg. The remaining $132$ subjects were born at term to uncomplicated pregnancies. Short-axis CINE MRI stacks were acquired with a $1.5$-T Siemens Sonata scanner. All images have a $7$mm slice thickness and $3$mm inter-slice gap, the in-plane resolution is $1.43 \pm 0.29$ mm (min. $0.57$, max $2.17$). All cardiovascular MRI was prospectively ECG gated and acquired during end-expiration breath holding. LV slices and endocardial masks were resampled into a homogeneous in-plane resolution of 2\textit{mm}, which yield slice pixel size of $212 \times 212$. Left ventricular short-axis endocardial borders were manually contoured by an expert reader at ES and ED using Siemens analytic software (Argus, Siemens Medical Solutions, Germany). The dataset was randomly divided into training, validation and testing sets of sizes $194, 20$ and $20$, respectively.

%    \subsubsection{GenerationR dataset}
%    
%    The Generation-R dataset has a total of 400 healthy individuals 9 years of age (training=340, validation=40 and test=40) \cite{generation-r2016}. All the LV stacks were acquired with a 3.0-T GE Discovery MR750w scanner. Images were respiratory and ECG-monitored, captured in 8-16 breath holds with a slice thickness of 7mm, inter-slice gap of 1mm and in-plane resolution was 1.09x1.09 mm. This dataset has almost double number of samples than Preterm dataset and almost 9-times more than MICCAI dataset, which is a good test for data hungriness of the deep learning models.

%\subsection{TwinsUK dataset}

\section{Recurrent fully-convolutional networks}

The proposed recurrent fully-convolutional network (RFCN) is an extension of the architecture originally introduced in \cite{long2015} for predicting pixel-wise, dense outputs from arbitrarily-sized inputs. The main idea underlying FCNs is to extend a contracting path, in which a sequence of pooling operators progressively reduces the size of the network, by adding successive layers where pooling operators are replaced by upsampling operators. In this respect, our architecture is similar to U-net \cite{ronneberger2015} where the expanding path is characterised by a large number of feature channels allowing the network to propagate context information to higher resolution layers.    

 \begin{figure*}[tb]
	\begin{center}
		\includegraphics[width=0.7\textwidth, clip, trim= 0 3.53cm 7cm 0]{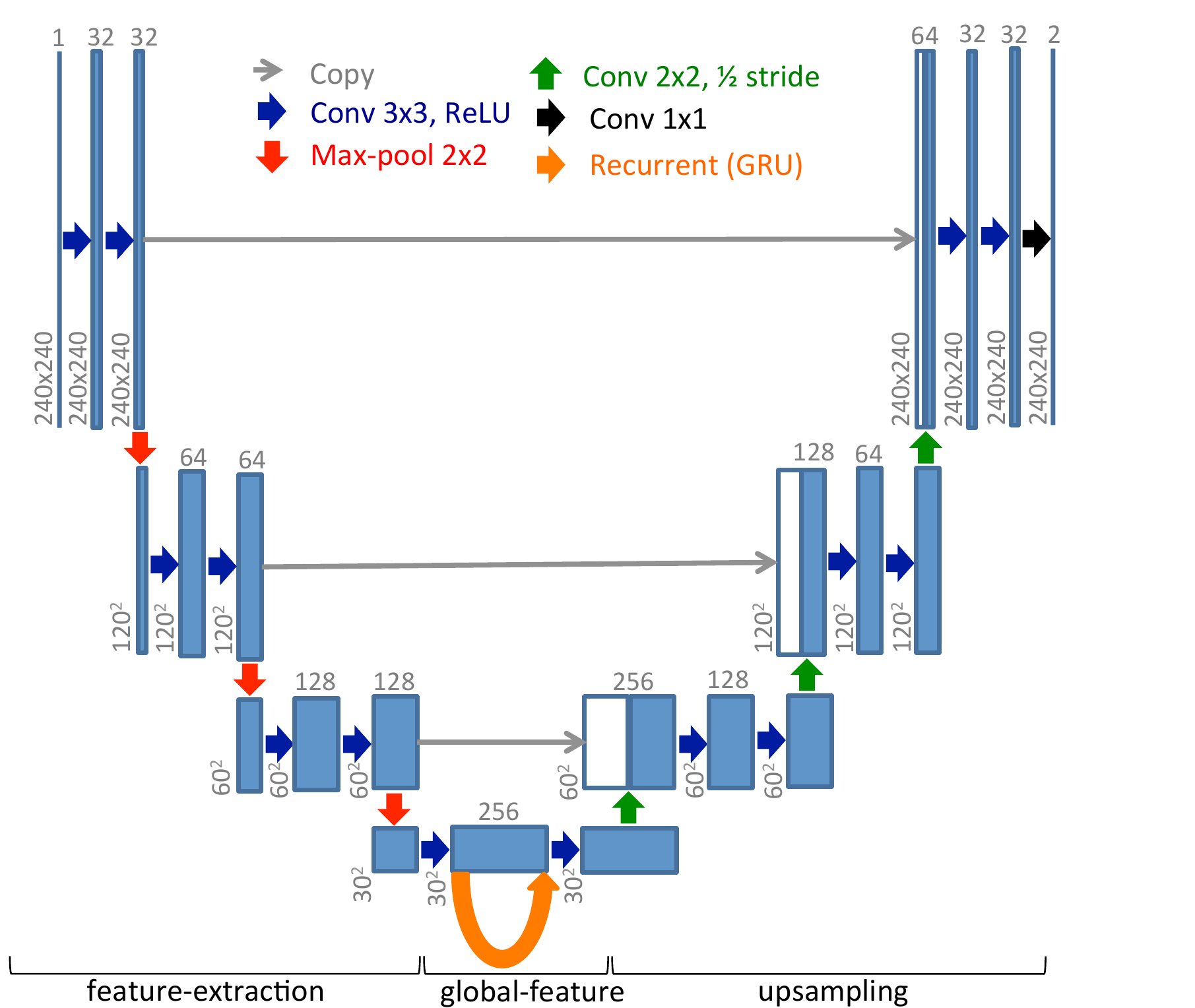}
	\end{center}
	\caption{Overview of the RFCN architecture. Blue boxes represent feature maps and white boxes represent copied feature maps. The number of feature maps and their dimensions are displayed above each box and on the left-side, respectively. Arrows represent network operations: gray arrows indicate copy operations, blue arrows indicate convolutional operations, red arrows indicate max-pooling operations, green arrows indicate convolutional operations with $1/2$ stride, black arrow indicates $1\times1$ convolutional operation and orange arrow indicates a recurrent connection to handle inter-slice dependences learned through GRU.}
	\label{fig:rfcn_model}
\end{figure*}

Our purpose is to model the full stack of short-axis images extracted from cardiac MRI and improve the segmentation of the left ventricle in each slice by leveraging inter-slice spatial dependences. The input is the entire sequence of $S$ slices obtained at a particular cardiac phase (ED or ES) and the output is the sequence of corresponding (manually produced) left-ventricular masks. Each input and output image is assumed to have equal size. A schematic illustration is given in Figure \ref{fig:rfcn_inputs}. As can be seen there, slices around the base of the heart (at the top) cover larger LV regions and show relatively clear boundaries whereas slices around the apex (at the bottom) cover smaller LV regions and present more blurred boundaries. Learning the typical shape deformations that are observed as we move from the base towards the apex is expected to improve the overall quality of the segmentation in challenging regions around the apex.    

Three main building blocks characterise the proposed RFCN as illustrated in Figure \ref{fig:rfcn_model}: a feature-extraction (contracting) path, a global-feature component and an upsampling (expanding) path. The feature-extraction component, which is independently applied to each image in the stack, deploys successive convolution and max-pooling operations to learn higher level features and remove local redundancy. In our architecture, this component consists of a repeated block of two $(3 \times 3)$ convolutional layers (with stride of $1$) followed by a rectified linear unit (ReLU) and a $(2 \times 2)$ max pooling layer (with stride of $2$). We doubled the number of feature channels $c$ after each max pooling layer to maintain enough context, i.e each block takes an input of size $(c \times h \times w)$ and generates output feature maps of size $(2c \times h/2 \times w/2)$.

At the end of this contracting path the network has extracted the most compressed features carrying global context. The global feature component starts here with a $(3 \times 3)$ convolutional layer (with stride of $1$) followed by a ReLU. We denote $\mathbf{e}_s$ the output of this layer where $s$ indicates the slice index, i.e. $s \in \{1, \ldots, S\}$. This output consists of $(256 \times 30 \times 30)$ feature maps. In an attempt to extract global features that capture the spatial changes observed when moving from the base to the apex of the heart, we introduce a recurrent mechanism mapping $\mathbf{e}_s$ into a new set of features, $\mathbf{h}_s = \phi(\mathbf{h}_{s-1}, \mathbf{e}_s)$, where $\phi(\cdot)$ is a non-linear function, and the size of $\mathbf{h}_{s}$ is the same as the size of $\mathbf{e}_s$. Another $(3 \times 3)$ convolutional layer (with stride of $1$) followed by a ReLU is then applied to complete the global-feature extraction block. 
Given that training recurrent architectures is particularly difficult due to the well-document vanishing gradient problem, several options were considered for the implementation of recurrent function $\phi$, including a Long Short-Term Memory (LSTM) \cite{hochreiter1997} and Gated Recurrent Units (GRUs) module \cite{cho2014}. GRUs in particular have been shown to achieve a performance comparable to LSTM on a number of tasks involving sequential data whilst requiring fewer parameters and less memory \cite{chung2014}. Here we have chosen to use a convolutional variant of GRU so that the local correlation of the input images are preserved whilst achieving a notable reduction in the number of parameters compared to its non-convolutional counter part.

For every slice, the dense feature maps that have been learned by the convolutional GRU module are then upsampled to compensate for the input size reduction caused by the max-pooling operations. The upsampled features are concatenated with a high resolution parallel layer aligned to the feature-extraction component, similarly to the U-net architecture \cite{ronneberger2015}. Our upsampling component consists of a repeated block of a convolutional layer (with a fractional stride of $1/2$), a feature map concatenation module and two $3 \times 3$ convolutional layers (with stride of 1) followed by ReLU. The feature map concatenation module combines the outputs of the upsample layer and parallel feature-extraction block. Each block of the upsampling component takes a three-dimensional input $c \times h \times w$ and output $c/2 \times 2h \times 2w$ dimensional tensor. A convolutional operation with fractional stride is employed to compensate the reduction in input size due to the max pooling operation. Even though the upsampling procedure smooths out the boundaries of the object to be segmented, the concatenation of up-sampled feature maps with high-resolution feature maps helps mitigate this smoothing problem by providing better local and boundary information. The final segmentation is obtained by using a $1 \times 1$ convolutional layer, which maps the output of the upsampling component onto the two classes, i.e. LV and background. The probability for each class is given by a softmax function across all pixel locations.

\section{Other architectures and model training}

Recently, deep belief networks (DBNs) have been proposed for automatic LV detection and segmentation using short-axis MR images \cite{ngo2013,ngo2014}. A DBN was first used to detect the region of interest containing the LV. Anatomical  segmentation was then carried out using distance-regularised level sets, which were modified to leverage prior shape information inferred by a separate DBN. In these models, as in FCNs, each slice in the short-axis stack is segmented independently of all the others. The main building block of a DBN model is a restricted Boltzmann machine (RBM), typically trained using the contrast divergence algorithm \cite{hinton2006}. In some of our experiments, we have assessed the performance of DBNs for LV segmentation comparably to FCNs and the proposed RFCNs.  

In order to further investigate whether modelling the dependence across slices typically yields improved performance, and motivated by the existing body of work on DBNs, we have also assessed the performance of a recurrent version of restricted Boltzmann machines (RRBM), originally proposed to learn human body motion \cite{sutskever2009}, but never used for LV segmentation. RRBMs are stacked together to form what we call a recurrent deep belief network (RDBN). Similarly to the proposed RFCN, RDBN takes the entire sequence of short-axis slices as input and leverages the spatial correlations through additional bias units. For further information, we refer the reader to the original work \cite{sutskever2009}.

%\begin{align*}
%J(\theta ) = - \frac{1}{N}\sum_{i=1}^{N}(ky^i \log(h_{\theta}(x^i)) + (1 - y^i) \log(1- h_{\theta}(x^i)))  
%\end{align*}
%where $\theta$ collects all the model parameters and $h$ indicate what model. In this notation, $N$ represents the total %number of pixels and $y$ is a binary indicator representing either the background ($y=0$) or the left-ventricle ($y=1$). %In both datasets, the number of background pixels largely outnumber the number of foreground pixels, hence we applied $k = %3.5$ times more weight for foreground. This weight can bias the model to pay more attention to the minority LV label. 
The two convolutional architectures, FCN and RFCN, were trained by minimizing the cross-entropy objective function. 
FCN was trained using a stochastic gradient descent algorithm with momentum whereas RFCN was trained using a stochastic gradient descent algorithm with RMSProp \cite{tieleman2012}. Back-propagation was used to compute the gradient of the cross-entropy objective function with respect to all parameters of the model, including the GRU component in the case of RFCN. We also learned $\mathbf{h}_0$ as required by the first slice of the sequence. In each block, batch normalization \cite{ioffe2015} was added after each convolutional layer, i.e. just before the max-pooling and upsampling layers. All reported results refer to the best out of $5$ experiments in which the models were initialised with random parameters. RFCN was initialised using weights obtained from FCN, which reduces the training time and provided the good initial weights. Both the DBN and RDBN architectures were trained using the contrast divergence algorithm \cite{hinton2006}. Dropout \cite{hinton2012} was found to improve their overall performance.
%For both DBN and RDBN, we used $1,000$ nodes in the first layer and $1,300$ nodes in second one. 
For all these models, best results were achieved using a learning rate of $0.01$ with constant decay of $3\%$ after each epoch, a momentum of $0.9$ and weight decay of $0.00005$. At the training phase, both the MICCAI and PRETERM datasets were augmented by generating additional artificial training images to prevent model overfitting. During training, we performed translation ($\pm 16$ pixels) and rotation ($\pm 40^{\circ}$) data augmentation, which was found to yield better performance. 

\section{Experimental results}

This section presents an empirical evaluation of several LV endocardium segmentation algorithms using three performance metrics:  good contours (GC) \cite{avendi2016}, Dice index, and average perpendicular distance (APD) between manually drawn and predicted contours \cite{radau2009}. In order to make our experimental results comparable with published studies on MICCAI dataset, all models were validated using the online set, and we report on results obtained on the validation set. Table \ref{tbl:results} summarises the experimental results. On the MICCAI dataset, the DBN-based results presented in \cite{ngo2014} include a Dice index of $0.88$, a GC of $95.71\%$ and an APD of $2.34$ mm whereas the pipeline described in \cite{avendi2016} results in a Dice index of $0.90$, a GC of $90\%$ and an APD of $2.84$ mm (before further post-processing). A comparable Dice index is obtained by both FCN and RFCN, which yield higher GC and smaller ADP. Here RFCN outperforms FCN with a substantially improved ADP of $2.05$ mm.  
%RFCN has more parameters than FCN, and a training sample of only $30$ scans may not be sufficient to handle the model complexity.

The PRETERM dataset was modelled using the the same architectures, without further customisation. The results of this application are also summarised in Table \ref{tbl:results}. For this dataset, we compared the performance of four different architectures: FCN, RFCN, DBN and RDBN. The latter two models were given as input a region of interest containing the LV thus conferring them an advantage compared to FCN and RFCN. On this dataset we were not able to test the recently proposed pipeline described in \cite{avendi2016}, which relies on multiple stages. As in the MICCAI dataset, the fully convolutional architectures have achieved superior performance. RFCN has outperformed all other architectures in terms of Dice index and APD, which was found to be as small as $1.56$ mm. In comparison, DBN with known LV location yields an APD of $2.05$ mm. 
 RDBN yields higher GC and lower APD compared to DBN thus providing additional evidence that performance gains can be obtained by modelling intra-slice dependences.

%%%%%%%%%%%%%%%%%%%%%%%%%%% MICCAI SUMMARY RESULTS 
%
%\begin{table}[t]
%\caption{MICCAI dataset: Summary of performance metrics.}
%\label{tbl:miccai_results}
%\begin{center}
%\begin{tabular}{llll}
% Model & Good contours (\%) & Dice Index & APD (mm) \\
%\hline \\
%Ngo et al. \cite{ngo2014} & $85.18~(15.83)$ & $0.850~(0.04)$ & $2.81~(0.47)$ \\
%Avendi et al. \cite{avendi2016} & $90.00~(10.00)$ & $0.900~(0.10)$ & $2.84~(0.29)$ \\ 
%FCN  		   & $94.78~(06.27)$ & $0.902~(0.04)$ & $2.14~(0.38)$ \\
%RFCN  		   & $95.34~(07.20)$ & $0.900~(0.04)$ & $2.05~(0.29)$ \\
%\hline
%\end{tabular}
%\end{center}
%\end{table}
%
%%%%%%%%%%%%%%%%%%%%%%%%%%% PRETERM SUMMARY RESULTS 
%
%\begin{table}[t]
%\caption{PRETERM dataset: summary of performance metrics. (\textsuperscript{**}) indicates that the true centre of the LV was manually determined.}
%\label{tbl:preterm_results}
%\begin{center}
%\begin{tabular}{llll}
% Model & Good contours (\%) & Dice Index & APD (mm) \\
%\hline \\
%DBN\textsuperscript{**} &  $92.01~(8.36)$ & $0.913~(0.02)$ & $2.05~(0.38)$ \\
%FCN     &  $97.59~(4.82)$ & $0.916~(0.03)$ & $1.80~(0.41)$ \\
%RDBN\textsuperscript{**}    &  $97.50~(6.77)$ & $0.909~(0.02)$ & $1.94~(0.23)$ \\
%RFCN    &  $95.37~(5.69)$ & $0.935~(0.03)$ & $1.56~(0.31)$ \\
%\hline
%\end{tabular}
%\end{center}
%\end{table}

%%%%%%%%%%%%%%%%%%%%%%%%%% SUMMARY RESULTS 
\begin{table*}[t]
\caption{Performance assessment on MICCAI and PRETERM datasets. For the MICCAI dataset, for completeness we also report on published results after a post-processing stage based on level-sets. The DBN and RDBN models only performed endocardium segmentation, not detection, i.e. they were applied to focused regions of interest centered around the left-ventricle. All other architectures performed LV detection and endocardium segmentation from full short-axis slices.}
\label{tbl:results}
\begin{center}
\begin{tabular}{lllll}
 Dataset & Model & GC  & Dice  & APD  \\
\hline \\
MICCAI with level-sets & \cite{ngo2014} & 95.91~(5.28) & 0.880~(0.03) & 2.34~(0.46) \\
& \cite{avendi2016} & 97.80~(4.70) & 0.94~(0.02) & 1.70~(0.37) \\ 
& \cite{ngo2017} & 95.91~(5.28) & 0.880~(0.03) & 2.34~(0.46) \\

\hline
\\
MICCAI without level-sets & \cite{ngo2014} & \textbf{95.71~(6.96)} & 0.880~(0.03) & 2.34~(0.45) \\
& \cite{avendi2016} & 90.00~(10.00) & 0.900~(0.10) & 2.84~(0.29) \\ 
& \cite{ngo2017} & 90.29~(12.73) & 0.880~(0.03) & 2.42~(0.36) \\
& FCN  		   & 94.78~(06.27) & \textbf{0.902~(0.04)} & 2.14~(0.38) \\
& RFCN  		   & 95.34~(07.20) & 0.900~(0.04) & \textbf{2.05~(0.29)} \\

\hline
\\
PRETERM  & DBN &  92.01~(8.36) & 0.913~(0.02) & 2.05~(0.38) \\
& RDBN  &  97.50~(6.77) & 0.909~(0.02) & 1.94~(0.23) \\
& FCN     &  \textbf{97.59~(4.82)} & 0.916~(0.03) & 1.80~(0.41) \\
&RFCN    &  95.37~(5.69) & \textbf{0.935~(0.03)} & \textbf{1.56~(0.31)} \\
\hline
\end{tabular}
\end{center}
\end{table*}

%%%%%%%%%%%%%%%%%%%%%%%%%% RESULTS BREAKDOWN

\begin{table*}
\caption{Breakdown of the Dice index by LV regions on both MICCAI and PRETERM datasets. Base-X and Apex-X represent number of slice(s) included starting from the base and apex of the left ventricle, respectively. The index is calculated using all slices from the all samples at once hence measuring overall pixels accuracy. DBN and RDBN performed the segmentation task using a pre-defined region of interest containing the LV region.}
\begin{center}
\begin{tabular}{l l c c c c c c c}
\multicolumn{1}{l}{Dataset}  &\multicolumn{1}{l}{Model} &\multicolumn{1}{l}{Base-1} &\multicolumn{1}{l}{Base-2} &\multicolumn{1}{l}{Base-3} &\multicolumn{1}{l}{Central} &\multicolumn{1}{l}{Apex-3} &\multicolumn{1}{l}{Apex-2} &\multicolumn{1}{l}{Apex-1} %&\multicolumn{1}{l}{Total}
\\ \hline \\
MICCAI &FCN &\textbf{0.9313} &\textbf{0.9314} &\textbf{0.9342} &0.9367 &0.8751 &0.8441 &0.7581 \\ %&0.9238

 &RFCN &0.9040 &0.9178 &0.9268 &\textbf{0.9433} &\textbf{0.9112} &\textbf{0.8917} &\textbf{0.8468} \\ %&\textbf{0.9307} \\
\hline
\\
PRETERM &DBN &0.9285 &0.9374 &0.9413 &0.9385 &0.8465 &0.7809 &0.6139 \\ %&0.9226

%DBN + LS \cite{ngo2013}\textsuperscript{*} &0.9526 &0.9556 &0.9548 &0.9220 &0.8936 &0.8558 &0.7232 &0.9324
%\\
%\hline \\
 &RDBN &0.9319 &0.9379 &0.9420 &0.9433 &0.8856 &0.8409 &0.7542 \\ %&0.9326

%RDBN + LS\textsuperscript{*} &0.9496 &0.9528 &0.9527 &0.9239 &0.8997 &0.8697 &\textbf{0.7980} &0.9334 
%\\
%BDBN &0.9387 &0.9431 &0.9450 &\textbf{0.9447} &0.8943 &0.8595 &\textbf{0.8046} &\textbf{0.9364}
%\\
 &FCN &0.9486 &0.9536 &0.9559 &0.9610 &0.9051 &0.8686 &0.7468 %&0.9490
\\
 &RFCN &\textbf{0.9576} &\textbf{0.9621} &\textbf{0.9631} &\textbf{0.9625} &\textbf{0.9178} &\textbf{0.8800} &\textbf{0.7571} \\ % &\textbf{0.9552} \\
 \hline
\end{tabular}
\end{center}
\label{tbl:results_break_down}
\end{table*}

In order to shed insights into the regional improvements introduced by RFCN, the Dice index was computed separately for different local regions of the LV, and the results are summarised in Table \ref{tbl:results_break_down}. Here, Base-$1$, Base-$2$ and Base-$3$ indicates that $1$, $2$ and $3$ slices were taken starting from the base of the heart and moving towards the middle, and analogously for the apex. All the remaining slices contributed towards the Central class. In all cases, the Dice index is calculated using all the samples at once to reflect overall pixels accuracy. In both datasets, RFCN outperforms FCN around the central slices and around the apex, as expected. However, in the MICCAI dataset, FCN yields better performance around the base of the heart. On the PRETERM dataset, both DBN and RDBN gave the worst performance, compared to FCN and RFCN, despite using focused region of interests instead of full-sized images. Here again it can be observed that RDBN improves upon DBN across all cardiac locations.

\begin{figure*}[tb]
	\begin{center}
		\includegraphics[width=\textwidth, clip, trim= 0.35cm 8.95cm 3.9cm 0]{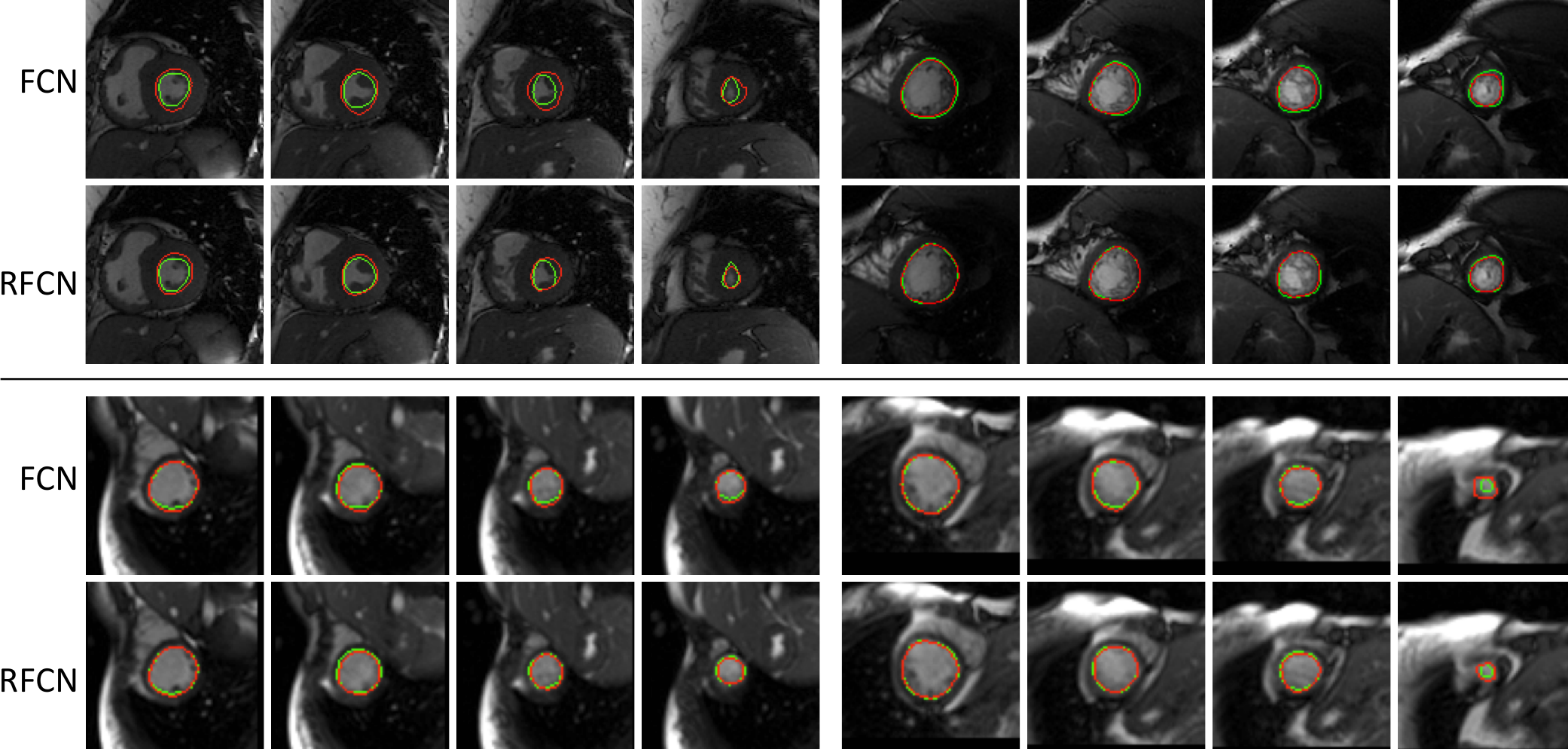}	
	\end{center}
	\caption{Example of segmented left ventricle using RFCN and FCN architectures from MICCAI dataset (top two rows) \cite{radau2009} and PRETERM dataset (bottom two rows). Green contours represent the ground truth and red contours are the predicted contours. RFCN is often able to better delineate the left-ventricle contours with weaker boundaries compared to FCN.}
	\label{fig:fcn_vs_rfcn}
\end{figure*}

\section{Conclusions}

%Recurrent neural network (RNN) and its variants are originally designed for one-dimensional sequence processing. Later, Graves et al. \cite{graves2007} proposed multi-dimensional LSTM network (MD-LSTM) for multi-dimensional data structure. MD-LSTM connect hidden LSTM units in grid like fashion, for example in two dimension image a hidden LSTM unit connect four LSTMs from up, down, left and right directions. Similarly, MD-LSTM can be use to 3D volumetric data source, however it would need 8 different LSTMs. Further, due to the sequential nature of MD-LSTM model, parallelization was difficult. Recently, Stollenga et al. \cite{stollenga2015} proposed efficient and easy to parallelize multi-dimensional LSTM model called PyraMiD-LSTM for volumetric image segmentation. In general, for $d$ dimensional source, PyraMiD-LSTM reduced number of required LSTM from $2^d$ in MD-LSTM to $2 \times d$ LSTMs. However, PyraMiD-LSTM is use as a network layer and uses fully-connected layers for classification. Hence, shares the disadvantages of sliding-window approach. Also, unlike in the multi-slice MR images, 3D shape of the objects has smooth transition between different plane in volumetric images. Hence,

In this paper we have investigated whether a single neural network architecture, trained end-to-end, can deliver a fully-automated and accurate segmentation of the left ventricle using a stack of MR short-axis images. The proposed architecture, RFCN, learns image features that are important for the localisation of the LV in a sequential manner, going from the base to the apex of the heart, through a recurrent modification of fully convolutional networks. 

Experimental findings obtained from two independent applications demonstrate that propagating information from adjacent slices can help extract improved context information with positive effect on the resulting segmentation quality. The hypothetical value of the large inter-slice correlation has been further tested by introducing a recurrent version of deep belief networks, and verified with our results showing that RDBNs generally outperform DBNs on the segmentation task, assuming the LV has already been localised. As expected, notable improvements can be seen in the delineation of cardiac contours around the apex, which are notoriously more difficult to identify. 

One surprising finding was to note that performance of RFCN in apical slices was better for MICCAI than for PRETERM cohort ($0.85$ vs. $0.76$ Dice index in the most apical slice, see Table 2), when one could expect the opposite: a regular and homogeneous cohort, PRETERM, should lead to a better performance when leveraging the inter-slice spatial dependence. This aspect will warrant further investigations. 

Compared to other models, RFCN has the advantage of carrying out both LV detection and segmentation in a single architecture with clear computational benefits and the potential for real-time application. In future work, we are planning to investigate alternatives operations that can capture inter-slice correlations, such as 3D convolutions, and further extend RFCN by incorporating a bi-directional mechanism for the inclusion of an inverse path (from the apex to the base of the heart) as well as a temporal extension to handle all cardiac phases at once.  

\section*{Acknowledgements}
The authors would like to thank Paul Leeson and Adam Lewandowski from Oxford University for their assistance with the PRETERM dataset.

%------------------------------------------------------------------------
% Bibliography
%
\bibliography{refs}

\begin{thebibliography}{10}
\providecommand{\url}[1]{\texttt{#1}}
\providecommand{\urlprefix}{URL }

\bibitem{avendi2016}
Avendi, M.R., Kheradvar, A., Jafarkhani, H.: A combined deep-learning and
  deformable-model approach to fully automatic segmentation of the left
  ventricle in cardiac {MRI}. Medical Image Analysis  30,  108--119 (May 2016)

\bibitem{cho2014}
Cho, K., van Merrienboer, B., Gulcehre, C., Bahdanau, D., Bougares, F.,
  Schwenk, H., Bengio, Y.: Learning {Phrase} {Representations} using {RNN}
  {Encoder}-{Decoder} for {Statistical} {Machine} {Translation}.
  arXiv:1406.1078  (2014)

\bibitem{chung2014}
Chung, J., Gulcehre, C., Cho, K., Bengio, Y.: Empirical {Evaluation} of {Gated}
  {Recurrent} {Neural} {Networks} on {Sequence} {Modeling}. arXiv:1412.3555
  (2014)

\bibitem{georgescu2005}
Georgescu, B., Zhou, X.S., Comaniciu, D., Gupta, A.: Database-guided
  segmentation of anatomical structures with complex appearance. In: {CVPR}.
  vol.~2, pp. 429--436 (2005)

\bibitem{hinton2006}
Hinton, G.E., Salakhutdinov, R.R.: Reducing the {Dimensionality} of {Data} with
  {Neural} {Networks}. Science  313(5786),  504--507 (2006)

\bibitem{hinton2012}
Hinton, G.E., Srivastava, N., Krizhevsky, A., Sutskever, I., Salakhutdinov,
  R.R.: Improving neural networks by preventing co-adaptation of feature
  detectors. arXiv:1207.0580 [cs]  (2012)

\bibitem{hochreiter1997}
Hochreiter, S., Schmidhuber, J.: Long {Short}-{Term} {Memory}. Neural
  Computation  9(8),  1735--1780 (1997)

\bibitem{hu2013}
Hu, H., Liu, H., Gao, Z., Huang, L.: Hybrid segmentation of left ventricle in
  cardiac {MRI} using gaussian-mixture model and region restricted dynamic
  programming. Magnetic Resonance Imaging  31(4),  575--584 (2013)

\bibitem{huang2004}
Huang, R., Pavlovic, V., Metaxas, D.N.: A graphical model framework for
  coupling {MRFs} and deformable models. vol.~2, pp. 739--746 (Jun 2004)

\bibitem{huang2011}
Huang, S., Liu, J., Lee, L.C., Venkatesh, S.K., Teo, L.L.S., Au, C., Nowinski,
  W.L.: An image-based comprehensive approach for automatic segmentation of
  left ventricle from cardiac short axis cine {MR} images. Journal of Digital
  Imaging  24(4),  598--608 (2011)

\bibitem{ioffe2015}
Ioffe, S., Szegedy, C.: Batch {Normalization}: {Accelerating} {Deep} {Network}
  {Training} by {Reducing} {Internal} {Covariate} {Shift}. arXiv:1502.03167
  [cs]  (Feb 2015)

\bibitem{jolly2009}
Jolly, M.: Fully {Automatic} {Left} {Ventricle} {Segmentation} in {Cardiac}
  {Cine} {MR} {Images} {Using} {Registration} and {Minimum} {Surfaces}. MIDAS
  Journal  49 (2009)

\bibitem{kass1988}
Kass, M., Witkin, A., Terzopoulos, D.: Snakes: {Active} contour models.
  International Journal of Computer Vision  1(4),  321--331 (1988)

\bibitem{lewandowski2013}
Lewandowski, A.J., Augustine, D., Lamata, P., Davis, E.F., Lazdam, M., Francis,
  J., McCormick, K., Wilkinson, A.R., Singhal, A., Lucas, A., Smith, N.P.,
  Neubauer, S., Leeson, P.: Preterm heart in adult life: cardiovascular
  magnetic resonance reveals distinct differences in left ventricular mass,
  geometry, and function. Circulation  127(2),  197--206 (2013)

\bibitem{li2010}
Li, C., Xu, C., Gui, C., Fox, M.: Distance {Regularized} {Level} {Set}
  {Evolution} and {Its} {Application} to {Image} {Segmentation}. IEEE
  Transactions on Image Processing  19(12),  3243--3254 (2010)

\bibitem{long2015}
Long, J., Shelhamer, E., Darrell, T.: Fully {Convolutional} {Networks} for
  {Semantic} {Segmentation}. In: {CVPR} (2015)

\bibitem{ngo2014}
Ngo, T.A., Carneiro, G.: Fully {Automated} {Non}-rigid {Segmentation} with
  {Distance} {Regularized} {Level} {Set} {Evolution} {Initialized} and
  {Constrained} by {Deep}-{Structured} {Inference}. In: {CVPR}. pp. 3118--3125
  (2014)

\bibitem{ngo2013}
Ngo, T.A., Carneiro, G.: Left ventricle segmentation from cardiac {MRI}
  combining level set methods with deep belief networks. In: {ICIP}. pp.
  695--699 (2013)

\bibitem{ngo2017}
Ngo, T.A., Lu, Z., Carneiro, G.: Combining deep learning and level set for the
  automated segmentation of the left ventricle of the heart from cardiac cine
  magnetic resonance. Medical Image Analysis  35,  159--171 (2017)

\bibitem{petitjean2011}
Petitjean, C., Dacher, J.N.: A review of segmentation methods in short axis
  cardiac {MR} images. Medical Image Analysis  15(2),  169--184 (2011)

\bibitem{radau2009}
Radau, P., Lu, Y., Connelly, K., Paul, G., Dick, A.J., Wright, G.A.: Evaluation
  {Framework} for {Algorithms} {Segmenting} {Short} {Axis} {Cardiac} {MRI}. The
  MIDAS Journal – Cardiac MR Left Ventricle Segmentation Challenge  (2009)

\bibitem{ronneberger2015}
Ronneberger, O., Fischer, P., Brox, T.: U-{Net}: {Convolutional} {Networks} for
  {Biomedical} {Image} {Segmentation}. In: MICCAI (2015)

\bibitem{sutskever2009}
Sutskever, I., Hinton, G.E., Taylor, G.W.: The {Recurrent} {Temporal}
  {Restricted} {Boltzmann} {Machine}. In: NIPS, pp. 1601--1608 (2009)

\bibitem{tieleman2012}
Tieleman, T., Hinton, G.: Lecture 6.5-rmsprop: Divide the gradient by a running
  average of its recent magnitude. COURSERA: Neural Networks for Machine
  Learning  4 (2012)

\bibitem{valipour2016}
Valipour, S., Siam, M., Jagersand, M., Ray, N.: Recurrent {Fully}
  {Convolutional} {Networks} for {Video} {Segmentation}. arXiv:1606.00487 [cs]
  (2016)

\end{thebibliography}
\bibliographystyle{splncs03}

\end{document}